\documentclass[letterpaper]{article}

\usepackage{amsmath}
\usepackage{natbib,alifeconf}  
\usepackage{caption,subcaption}
\usepackage{hyperref}

\title{Growing Isotropic Neural Cellular Automata}

\author {Alexander Mordvintsev,$^{1}$ Ettore Randazzo,$^{1}$ \and Craig Fouts$^{2}$\\
\mbox{} \\
$^1$ Google Research \\
\{moralex, etr\}@google.com \\
\\
$^2$ The Ohio State University\\
fouts.57@osu.edu}

\begin{document}
\maketitle

\begin{abstract}

Modeling the ability of multicellular organisms to build and maintain their bodies through local interactions between individual cells (morphogenesis) is a long-standing challenge of developmental biology. Recently, the Neural Cellular Automata (NCA) model was proposed as a way to find local system rules that produce a desired global behaviour, such as growing and persisting a predefined target pattern, by repeatedly applying the same rule over a grid starting from a single cell. In this work, we argue that the original Growing NCA model has an important limitation: anisotropy of the learned update rule. This implies the presence of an external factor that orients the cells in a particular direction. In other words, “physical” rules of the underlying system are not invariant to rotation, thus prohibiting the existence of differently oriented instances of the target pattern on the same grid. We propose a modified Isotropic NCA (IsoNCA) model that does not have this limitation. We demonstrate that such cell systems can be trained to grow accurate asymmetrical patterns through either of two methods: \textbf{(1)} by breaking symmetries using structured seeds or \textbf{(2)} by introducing a rotation-reflection invariant training objective and relying on symmetry-breaking caused by asynchronous cell updates.

\end{abstract}

\section{Introduction}

Every multicellular organism begins its life as a single cell. Descendants of this egg cell reliably form complex structures of an organism through a process of division and differentiation, also known as morphogenesis \citep{Turing1990-yu}. In many cases, this process doesn't require any external control or orchestration and is described as self-organizing; cells communicate with their neighbors to make collective decisions about the overall body layout and composition. Understanding this process is an active area of research \citep{Pezzulo2016-ke} with a number of models proposed to explain the development procedure of various tissues \citep{Malheiros2020} and organisms.

There is a wide spectrum of existing models that either attempt to reproduce biological processes or simplify them in order to accomplish engineering tasks or simulate artificial life. One widespread approach is the use of artificial Gene Regulatory Networks (GRN) \citep{Cussat2019,Jong2002}, which have, for instance, been used to model cell growth \citep{10.1162/ARTL_a_00075, Jacobsen1998-sl}. Discovering effective models with these networks often relies on manual engineering or genetic algorithms. Although these models have demonstrated impressive results, we believe that, in a plausible time and compute budget, genetic algorithms are limited in the complexity and fitness accuracy they can achieve. Differentiable programming has been demonstrated as a powerful and versatile strategy for solving complex engineering problems, including morphogenesis modeling.

One recent differentiable approach to modelling morphogenesis is based on Neural Cellular Automata (NCA) \citep{mordvintsev2020growing}. Here, the authors represent a growing organism with a uniform grid of raster cells where the state of each cell is characterized by a set of scalar values. Cells repeatedly update their states using a rule defined by a small neural network that takes as input information collected from each cell's neighbors at the current moment in time. Backpropagation through time is used to learn the local update rule that satisfies the global objective of growing a predefined target pattern, allowing for the discovery of much more complex rules than what has been possible with existing strategies. While this is an attractive direction to pursue for modeling complex systems, we observe that this model is still not \textit{fully} self-organising.

\subsection{Anisotropy of Neural CA}

\begin{figure}[h]
  \includegraphics[width=\columnwidth]{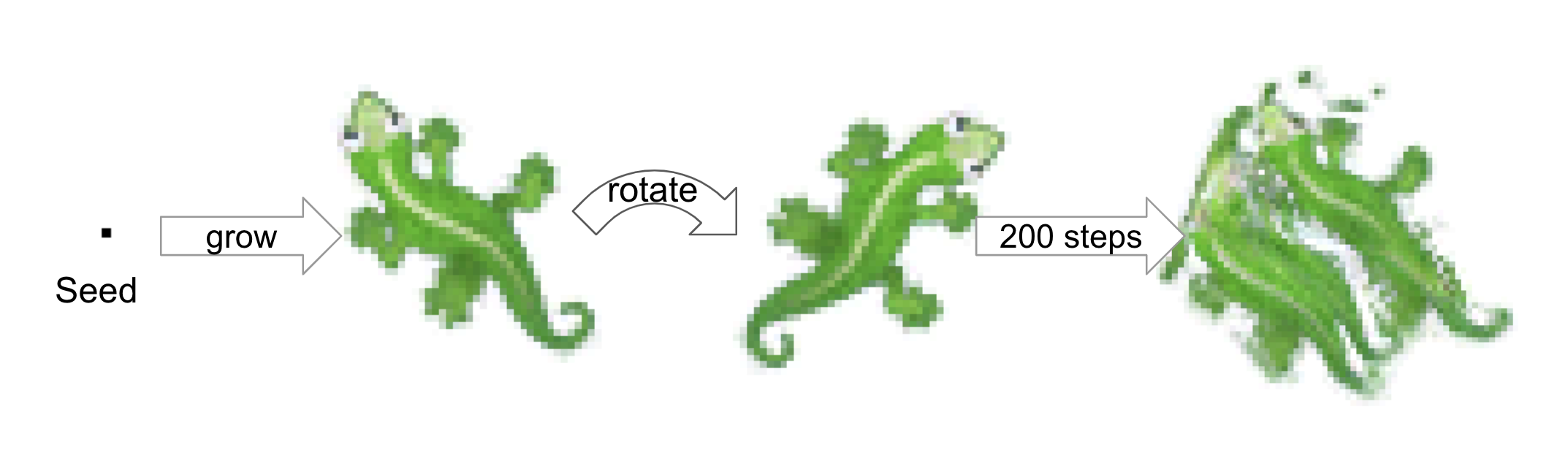}
  \caption{Anistropy of the Growing NCA model. Cells rely on externally-provided global cell alignment and are unable to sustain a pattern if cell states are re-sampled in a rotated coordinate frame. In contrast, real living creatures can usually tolerate rotation without exploding.}
  \label{fig:aniso_problem}
\end{figure}

Figure \ref{fig:aniso_problem} demonstrates the weakness of the original Growing NCA model which challenges the claims of fully self-organizing pattern growth achieved by this model. This model can only grow and persist patterns in a specific orientation that is determined by properties of the space itself rather than the intrinsic states of the cells occupying said space. The NCA anisotropy stems from the axis-aligned Sobel filters that are used to model cell perception. In the last experiment, \footnote{\url{https://distill.pub/2020/growing-ca/\#experiment-4}} authors show that altering properties of the grid (Sobel filter directions) leads to rotations of the resultant pattern, but don't address the main concern that pattern orientation should be defined by the configuration of cells occupying the space and not a property of the space itself. In the follow up work on NCA texture synthesis \citep{niklasson2021self-organising}, the same group of authors experiment with varying the filter directions across space, reinforcing the idea of external control on each cell's perception.

In this work we argue that the original Growing NCA model is not \emph{fully} self-organizing due to a limitation in the model's architecture. The learned update rule is anisotropic, which implies the presence of an external factor that orients all cells in a particular direction. In other words, “physical” rules of the underlying system are not invariant to rotation, prohibiting the formation of differently oriented instances of the target pattern on the same grid. This would be akin to an animal only able to grow, or even exist, when facing north. We aim to relax this limitation with following contributions:

\begin{itemize}
    \item Propose a simplification to the original NCA update rule to make it isotropic, so the perception of each cell is invariant to rotation or reflection of the grid.
    \item Show that this invariance enables us to perform rotations, reflections, and other augmentations on structured seeds that predictably influence the model's behavior. 
    \item Design a rotation-reflection invariant training objective that steers the system towards reliable growth of asymmetric, anisotropic patterns through symmetry-breaking rather then external guidance.
    \item Demonstrate the robustness of the learned NCA rule to out-of-training grid structures.
\end{itemize}

\section{Isotropic Neural CA model}

The Isotropic NCA (IsoNCA) model described here can be seen as a more restricted version of the Growing NCA model, where the Sobel X and Y perception filters are replaced with a single Laplacian filter. This section covers key features of the model design.

\paragraph{Grid} Cells exist on a regular Cartesian grid; the state of each cell is represented by a vector $$\mathbf{s} = [s^0=R, s^1=G, s^2=B, s^3=A, ... , s^{C-1}]  \label{eq:s}$$ where $C$ is the number of channels and the first four channels represent a visible RGBA image. Initially, the whole grid is set to zeros, except the seed cell in which $A=1$. Cells iteratively update their states using only the information collected from their 3x3 Moore neighbourhood.

\paragraph{Stochastic updates} Cell updates happen stochastically; at every NCA step each cell is updated with probability $p_\text{upd}$ (we use value 0.5 in our experiments). This stochasticity is meant to eliminate dependence on a global shared clock that synchronizes the updates between cells. Previous work on NCA discusses the impact of this strategy on NCA robustness \citep{niklasson2021asynchronicity}. In IsoNCA models, asynchronicity plays a critical role in the symmetry breaking process (see the Results section). This asynchronicity can be seen as a strategy for generating noise, which has been documented to help biological systems construct complex functions in simple ways \citep{Samoilov2006-xj}.

\paragraph{``Alive'' and ``empty'' cells} The alpha channel ($s^3=A$) plays a special role in determining whether a cell is currently ``alive'' or ``empty''; each cell is alive if $A>0.1$ or if it has at least one alive cell in its 3x3 neighbourhood. The state of empty cells is explicitly set to zeros after each CA step.

\paragraph{Perception} Each cell collects information about the state of its neighborhood using a per-channel discrete 3x3 Laplacian filter. This filter computes the difference between the state of the cell and the average state of its neighbours. A cell's perception vector is the concatenation of its own current state and per-channel Laplacians of its neighbourhood: $$\mathbf{p} = concat(\mathbf{s}, K_{lap} \ast \mathbf{s})$$ where $\mathbf{s}$ denotes the cell's state and $K_{lap}$ is given by
$$
K_{lap} = 
\begin{bmatrix}
1 & 2 & 1\\2 & -12 & 2 \\1 & 2 & 1 \\
\end{bmatrix}
$$

\paragraph{Update rule} Cells stochastically update their states using a learned rule that is represented by a two-layer neural network: $$\mathbf{s}_{t+1} = \mathbf{s}_t + relu(\mathbf{p}_tW_0+b_0)W_1$$
where parameters $W_0$, $b_0$, and $W_1$ have shapes $(32, 192)$, $(192)$ and $(192, 16)$ respectively, which gives a total of 9408 learned parameters.

\section{Training IsoNCA}

The original Growing NCA model was trained to learn an update rule starting from a single seed cell and based on a target pattern fixed in a specific orientation. Under this training regime, our isotropic restrictions prevent the model from breaking spatial symmetries, demanding we either relax the target objective or further specify the initial conditions. Here, we propose and present implementations of both of these strategies.

\subsection{Structured seed strategy}

\begin{figure}[h]
    \centering
    \begin{subfigure}[b]{\columnwidth}
        \centering
        \includegraphics[width=\columnwidth]{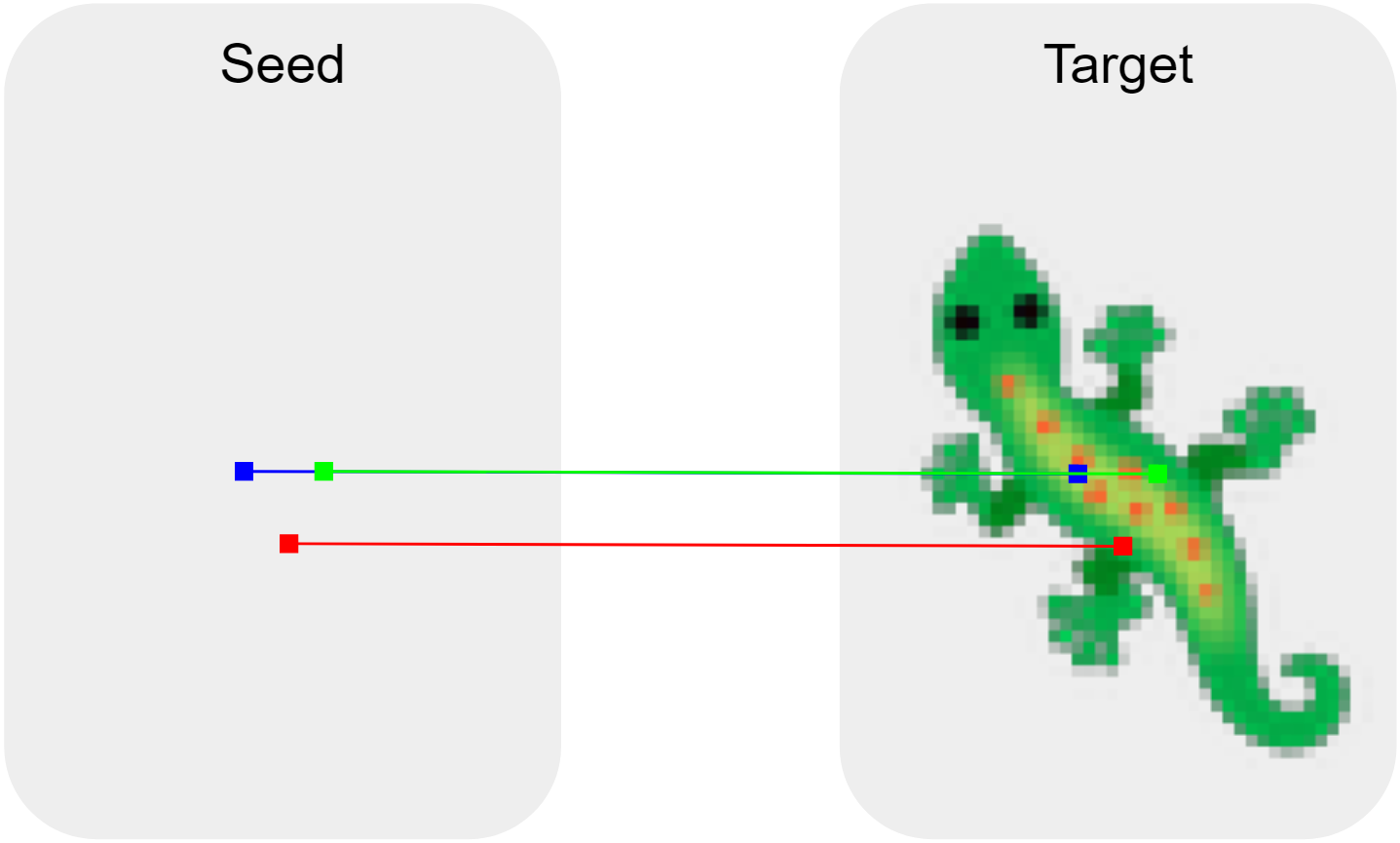}
        \caption{3-point rotation-reflection objective.}
        \label{fig:lizard_map}
        \vspace*{3mm}
    \end{subfigure}
    \begin{subfigure}[b]{\columnwidth}
        \centering
        \includegraphics[width=\columnwidth]{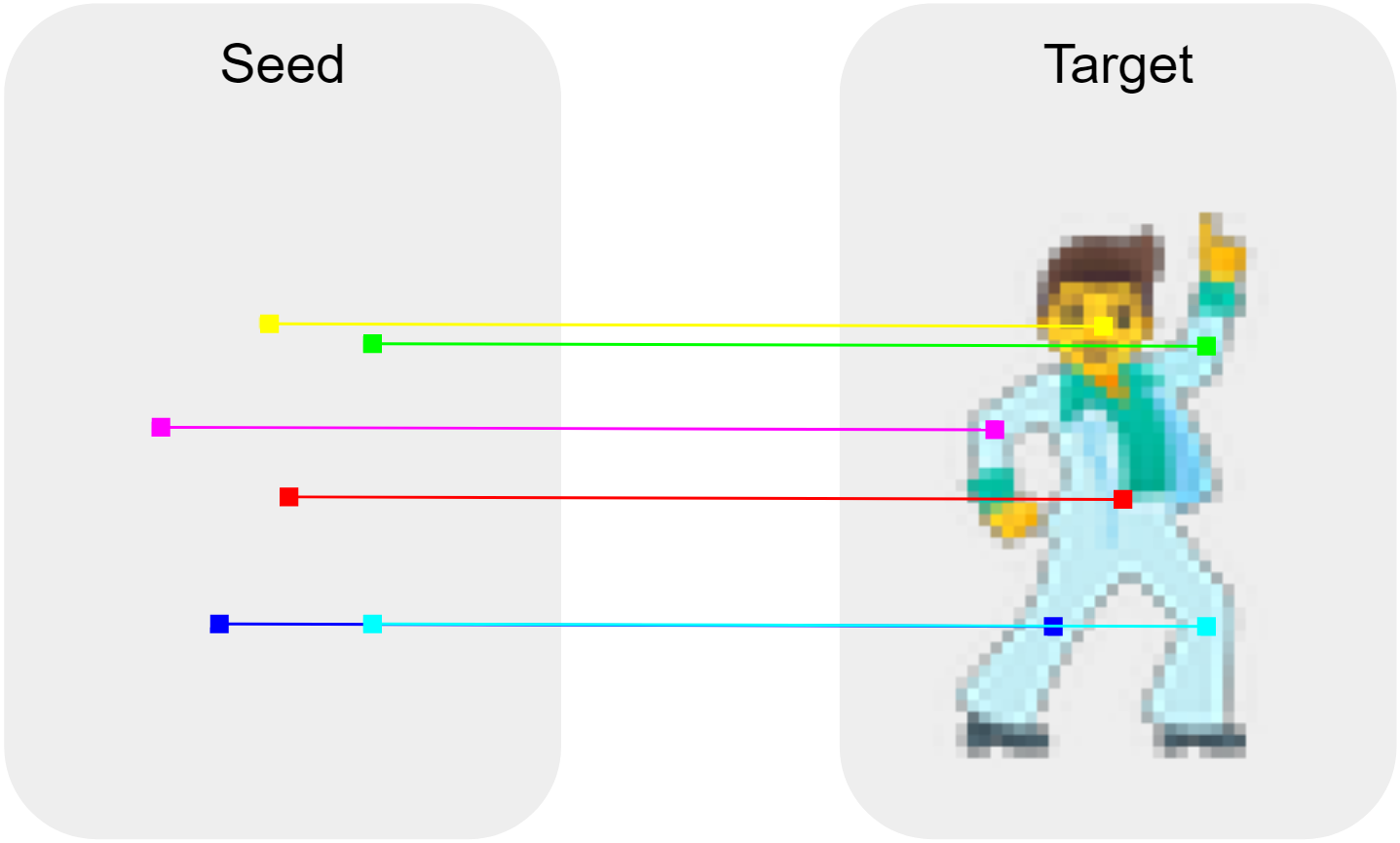}
        \caption{6-point structural mutation objective.}
        \label{fig:dancer_map}
    \end{subfigure}
        \caption{Mapping between structured seed points and their respective structural features in the target pattern. Points are placed manually to correspond with these features and distinct point colors are generated by converting equidistant HSV hues into RGB encodings.}
        \label{fig:target_maps}
\end{figure}

One way to train our isotropic model is by introducing a more comprehensive seed structure (e.g. by using three or more distinct points) in order to break symmetries of the system from the initial state. This is inspired by synthetic reaction-diffusion networks used to govern complex pattern growth \citep{scalise2014}. A \textit{structured seed} is defined by its number of points, the initial channel encodings of those points, and their positions relative to one another. Given that an isometry relating any pair of congruent triangles is unique \citep{coxeter1961}, we use 3 non-collinear points to define the seed's orientation, with points distributed uniformly on a circular edge of predefined radius. To establish directional responsibility, points are distinguished by their RGB encodings. An example of how one such seed maps to a target pattern is shown in Figure \ref{fig:lizard_map}. Training our model to grow in alignment with this seed then proceeds identically to the training of the original Growing NCA's single-seed scenario.

Alternatively, structured seeds can be manually engineered to map key features of the target pattern to specific points of the seed. So long as there are three or more non-collinear points, this configuration enables the model to break symmetries similarly to the triangular seed. Figure \ref{fig:dancer_map} shows one such mapping between the appendages of a dancer pattern and their respective point assignments in the corresponding seed. These points are reconfigurable and consequently used to grow predictable out-of-training structural mutations of the target pattern. Changing the configuration of a structured seed is performed by modifying the positions and channel encodings of its composite points. For example, if a structured seed is comprised of points $A$ and $B$, then replacing point $A$ with point $B$ involves adjusting the RGB encoding of point $A$ to match that of point $B$. 

\subsection{Single-seed strategy}

\begin{figure*}[ht]
  \includegraphics[width=\textwidth]{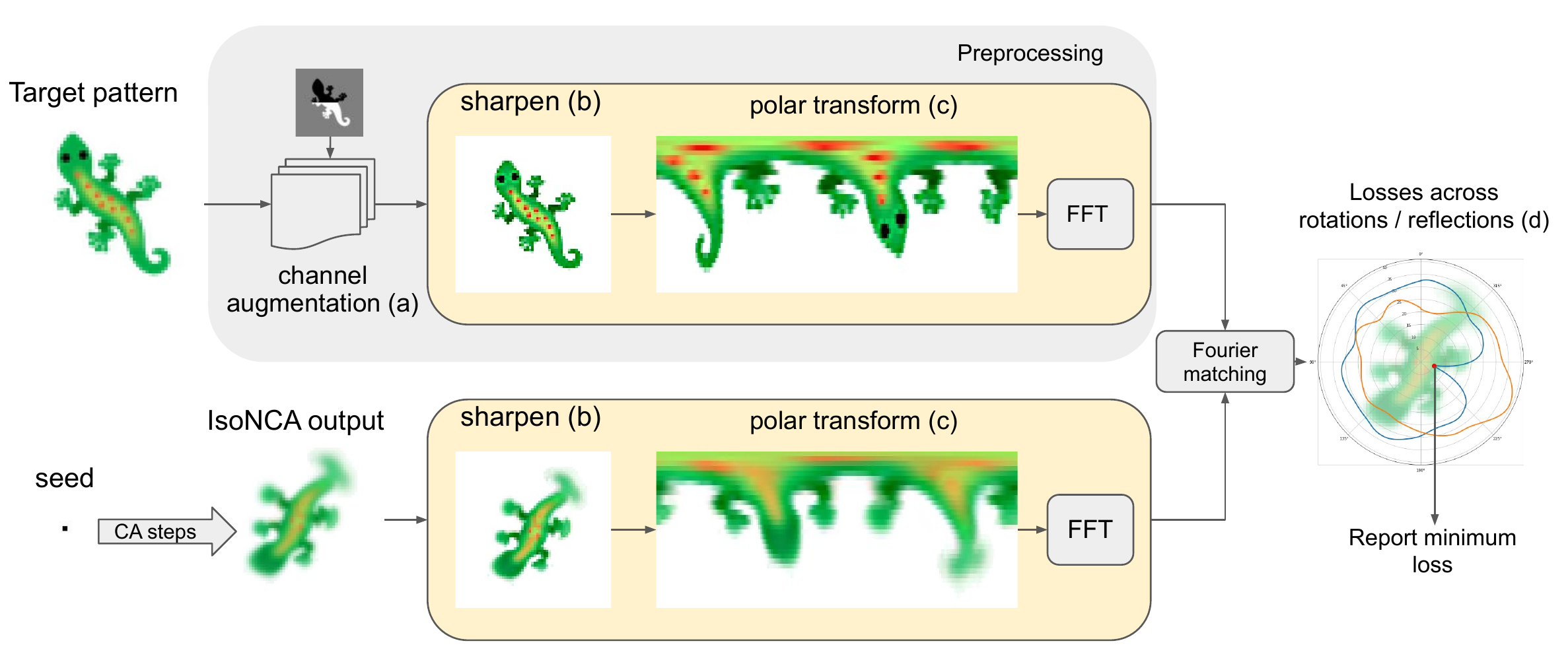}
  \caption{Rotation / reflection-invariant IsoNCA training pipeline. The target pattern is augmented with extra channels \textbf{(a)} to break symmetries that may interfere with training. Both the target and NCA-grown patterns are sharpened \textbf{(b)} to steer optimization into preserving fine details. A polar transformation \textbf{(c)} is applied to turn the unknown rotation between two images into a horizontal shift. Fourier-domain image matching enables efficient computation of pixel-matching losses across all orientations and reflections \textbf{(d)}. The blue plot shows losses with respect to the original pattern, and the yellow with respect to its reflected version.  The minimal loss value is selected for backpropagation.}
  \label{fig:invariant_loss}
\end{figure*}

Similar to the Growing NCA work, we also train NCA models that grow and persist a predefined pattern on a plane starting from a single seed cell. We use pixel-wise differences to match the pattern produced by the trained model to the target, but modify the loss function to make it rotation-reflection invariant. Due to the rotation symmetry of the cell perception we are unable (and do not want) to enforce a particular pattern orientation and chirality during training. Instead we select an individual rotation and reflection of the target that minimizes the pixel-wise loss value for each NCA-generated sample in the training batch. We expect this to cause the model to amplify noise from the stochastic cell updates and break the rotational symmetry to make a collective decision on the grown pattern orientation.

A naive implementation of the previously described procedure would require matching against densely sampled rotated and reflected instances of the training pattern, which is computationally inefficient. Instead, we  use a polar coordinate transformation and FFT to efficiently compute the discrepancy across different target rotations. Figure \ref{fig:invariant_loss} shows the architecture of the proposed loss function; here, we match patterns in the polar coordinate system, where rotations become horizontal translations. For each possible rotation angle $\theta$, we compute the sum of squared pixel-wise differences between the NCA-generated pattern $S$ and the target $T$ across all radius values $r$ and target channels $c$. Target channels include RGBA color representations and extra auxiliary channels that help the optimization break symmetries and escape sub-optimal local minima (see Figure \ref{fig:aux_losses}). The rotation-invariant loss can be expressed as follows:
\begin{gather*}
L(S, T) = \min_\theta \sum_{r,c} L_{r, c, \theta} \\ 
L_{r, c, \theta} = \sum_{\theta'} (S_{r, c, \theta'} - T_{r, c, \theta'-\theta})^2 = \\
\sum_{\theta'} S_{r, c, \theta'}^2 + \underbrace{\sum_{\theta'} T_{r, c, \theta'-\theta}^2}_\text{doesn't depend on $\theta$ } - 
2  \underbrace{ \sum_{\theta'} S_{r, c, \theta'} T_{r, c, \theta'-\theta} }_\text{1D convolution}
\end{gather*}
The 1D convolution term in this expression can be efficiently computed for all values of $\theta$ using the convolution theorem: $$s \ast t = \mathcal{F}^{-1}(\mathcal{F}(S)\cdot\mathcal{F}(T)),$$
where $\mathcal{F}$ denotes the Fourier transform and "$\cdot$" indicates elementwise multiplication. To make the pattern-matching loss reflection-invariant, we also compute the loss for a reflected version of the target pattern, and select the minimum between the two:  $$L_{\text{inv}} = \min(L(S, T), L(S, reflect(T)))$$

\subsubsection{Auxiliary channels}

\begin{figure}[h]
 \centering
  \includegraphics[width=0.9\columnwidth]{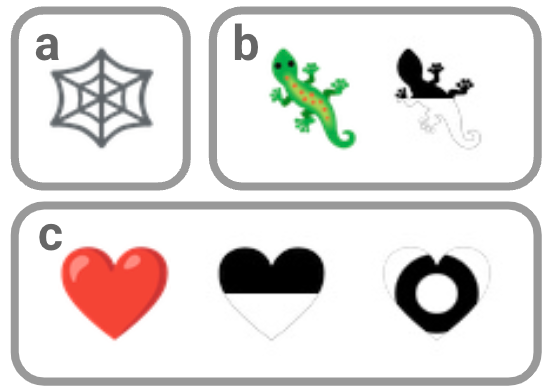}
  \caption{Target patterns and auxiliary targets used for the ration-reflection invariant training. \textbf{(a)} the spiderweb does not use auxiliary targets. \textbf{(b)} the lizard uses the binary auxiliary target. \textbf{(c)} the heart uses both binary and radial encoding auxiliary targets.}
  \label{fig:aux_losses}
\end{figure}

Having a rotation-invariant loss causes some patterns to exhibit strong local minima. For instance, models trained on the lizard pattern are be unable to reliably break one symmetry and often create a mixture of head-tails along with an assortment of limbs because they fail to differentiate up from down (see Figure \ref{fig:lizard_noaug}). This issue can be rectified using an auxiliary loss by adding a new ``target channel" to the rotation-reflection invariant loss in which the image is split in half and the upper and lower parts of the image have target values of $-0.5$ and $+0.5$ respectively (see Figure~\ref{fig:aux_losses}b). We call this additional channel the ``binary auxiliary channel." Including it enables models to learn to break symmetries and results in a much smoother loss. We have not observed any need to add an additional perpendicular auxiliary target to break the left-right symmetry.

We have also observed that IsoNCA struggles to generate uniform patterns, such as the heart emoji, suffering from stability issues and being unable to form the proper shape. We hypothesized that the shape of the pattern does not facilitate smooth training and decided to enhance the target image in a similar way to how we break symmetries using another auxiliary loss. Inspired by positional encodings used in the transformer architecture \citep{vaswani2017attention}, we add target channels generated by aliased concentric circles with the following rule: given a point with distance $r$ from the center and mode n, the positional encoding of the point is given by
$$
\begin{cases}
    sign(sin(r*n*\pi))*0.5,& \text{if n is even}\\
    sign(cos(r*n*\pi))*0.5,         & \text{otherwise}
\end{cases}
$$
Thus, the image can have arbitrarily many auxiliary channels with different modes to generate radial encodings. For the heart emoji, one radial encoding with mode 4 is sufficient (Figure~\ref{fig:aux_losses}c).

In this paper, we showcase three example patterns: the spiderweb, where no auxiliary loss is needed; the lizard, where the binary auxiliary channel is needed; and the heart, where both the binary and radial encoding auxiliary channels are needed (see Figure~\ref{fig:aux_losses}).

\section{Results}

\subsection{Structured seed experiments}

\begin{figure}[h!]
  \includegraphics[width=\columnwidth]{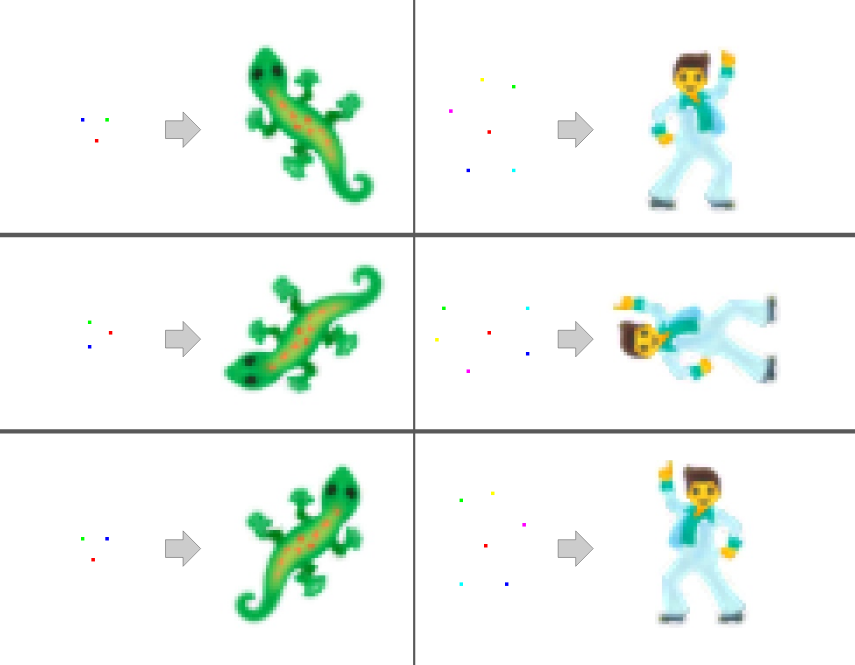}
  \caption{Pairs of initial seed configurations and their resulting unfoldings at step 5000 for several out-of-training isometries.}
  \label{fig:structured_results}
\end{figure}

\begin{figure}[h!]
  \includegraphics[width=\columnwidth]{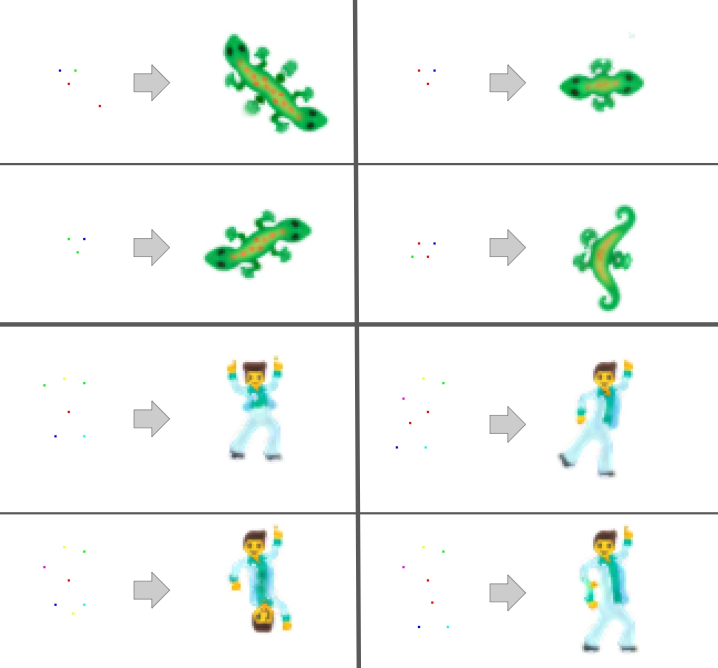}
  \caption{Pairs of initial seed configurations and their resulting unfoldings at step 5000 for several out-of-training mutations.}
  \label{fig:structured_out_of_training}
\end{figure}

In the original Growing NCA work, it is shown that the model can learn to grow at predefined angles, but it fails to extrapolate to unseen orientations after training is complete. Our model not only successfully addresses this limitation, but expands upon how orientation can be implicitly encoded using structured seeds.

Distributing growth responsibility between the points of a structured seed enables us to manipulate the model's behavior by performing plane isometries on the seed. In performing such an isometry, we see the transformation propagate throughout the model's growth of the resultant pattern without loss of stability. As shown in Figure \ref{fig:structured_results}, IsoNCA exhibits stable rotations and reflections irrespective of the fixed orientation of the target pattern.

Apart from plane isometries, we observe that the isotropic properties of our model emerge within sub-regions of the growing pattern. Reconfiguring the structured seed after training thus causes our model to manifest structural mutations corresponding to the modified seed configuration. As shown in Figure \ref{fig:structured_out_of_training}, our model grows a variety of out-of-training patterns exhibiting seed-directed mutations while maintaining the cohesion of the resultant shape. The stability of such irregular modifications is observed to be less reliable than that of plane isometries, however, and further seed manipulation often yields improved results. For example, by adjusting the supplanted seed point's relative position to more closely resemble that of its original counterpart, we often find that mutations stabilize with more consistently desirable behavior.

\subsection{Single seed experiments}

\begin{figure}[h]
  \includegraphics[width=0.45\columnwidth]{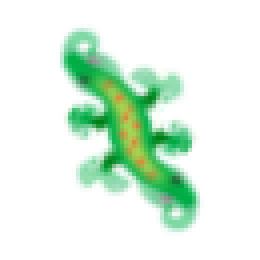}
  \includegraphics[width=0.45\columnwidth]{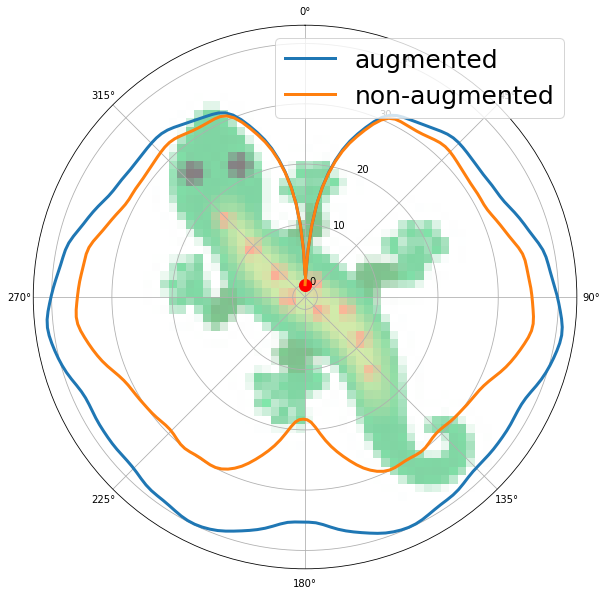}
  \caption{\textbf{Left:} the result of training IsoNCA to grow a non-augmented lizard pattern. Optimization struggles to break the head-tail symmetry. \textbf{Right:} matching losses computed between the target pattern and it's rotated instances. The non-augmented pattern has a strong local minima corresponding to a $180^\circ$ rotation. Augmenting the target with a non-symmetrical auxiliary channel flattens the spurious minima and facilitates convergence to the correct solution.}
  \label{fig:lizard_noaug}
\end{figure}

\paragraph{Target augmentation} As discussed in the "auxiliary channels" section, we augment some target patterns with extra channels to facilitate optimization convergence to a desirable solution. The lizard pattern is a particularly interesting case; without augmentation, the optimization fails to break the symmetry between the head and the tail, which leads to formation of the ghostly symmetrical pattern seen in Figure \ref{fig:lizard_noaug}, left. What's more, the heart pattern fails to converge to a definite shape at all without both radial and top-bottom contrast augmentations.

\begin{figure}[h]
  \includegraphics[width=\columnwidth]{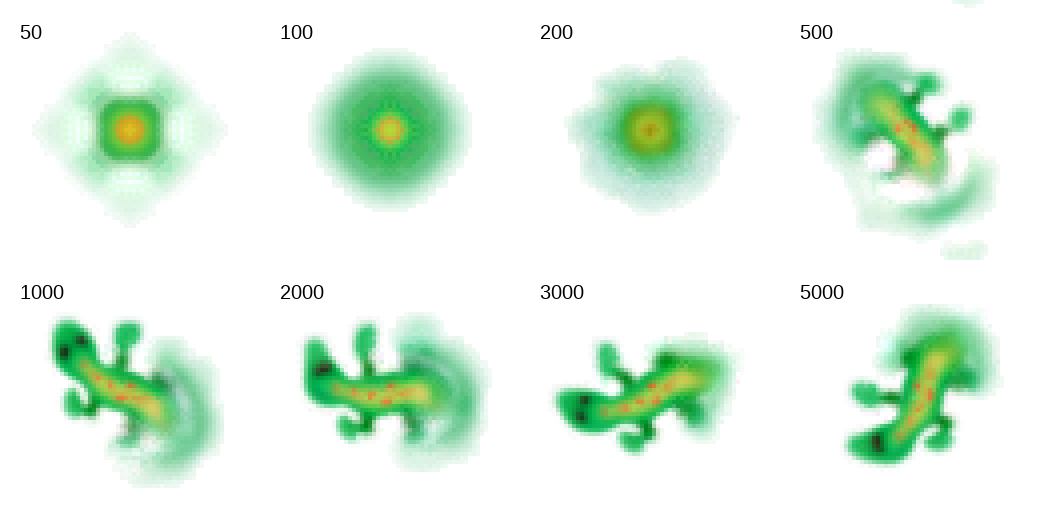}
  \hrule
  \includegraphics[width=\columnwidth]{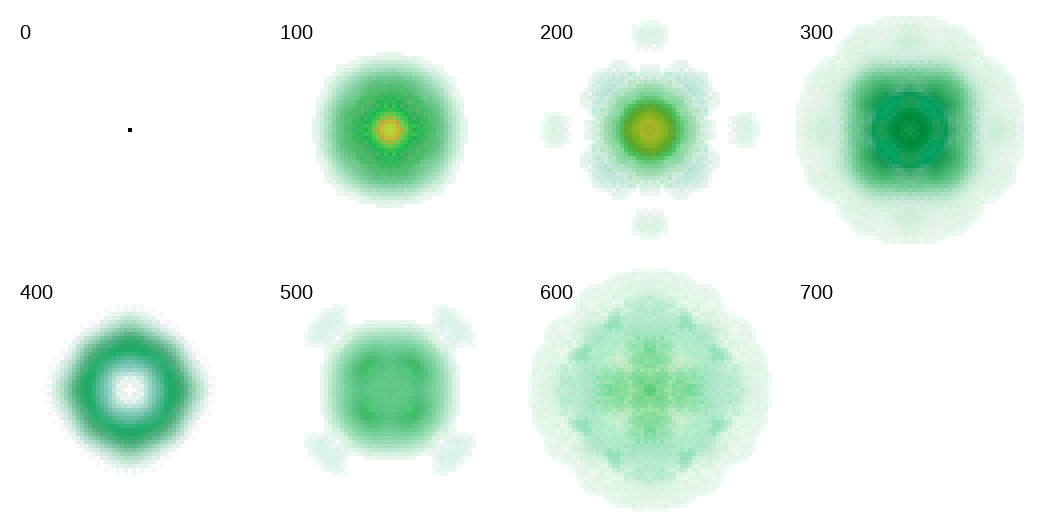}
  \caption{\textbf{Top:} IsoNCA manages to break the symmetry even in the case of deterministic synchronous cell updates by exploiting the non-associativity of floating-point number addition in the Laplacian filter convolution. 
  \textbf{Bottom:} the evolution of IsoNCA in the case of fully synchronous cell updates and perfectly symmetric perception. As expected, the model can't break symmetries to develop features of the lizard. The pattern produced by this particular checkpoint deterministically vanishes after about 700 steps. Other checkpoints may produce stable or exploding behaviours in this out-of-training regime.}
  \label{fig:sync_updates}
\end{figure}

\paragraph{Stochastic symmetry breaking} Growing non-rotationally-symmetrical patterns starting from a single seed is fundamentally different from the structured seed scenario. For example, on a regular square grid, both a cell's perception field and starting condition have symmetries that need to be broken during pattern development. We use stochastic asynchronous cell updates, which cells may rely on as a source of randomness. During training, cells successfully develop a protocol for making a collective decision about the final pattern layout that appears to rely on this randomness. To validate this assumption, we ran the IsoNCA rule, trained in the stochastic update regime, in a fully synchronous setting ($p_\text{upd}=1$). We expected that our model wouldn't be able to break symmetries between grid directions, and would produce some $90^{\circ}$ rotation- and reflection-symmetrical patterns. To our surprise, we instead observed that after approximately 100-150 iterations, asymmetries develop, and eventually our model produces an incomplete, unstable, but definitely not symmetrical version of the target pattern (see Figure \ref{fig:sync_updates}, top). Puzzled by this behaviour, we discovered that our model was able to exploit the non-associativity of floating point number accumulation in the convolution with the Laplacian filter to break the symmetry. We then implemented an associative version of the Laplacian filter by performing the convolution using fixed-point number representation. This time, as expected, the model produced a symmetrical pattern that vanished after oscillating for about 700 steps (see Figure \ref{fig:sync_updates}, bottom). 

\begin{figure}[h]
  \includegraphics[width=\columnwidth]{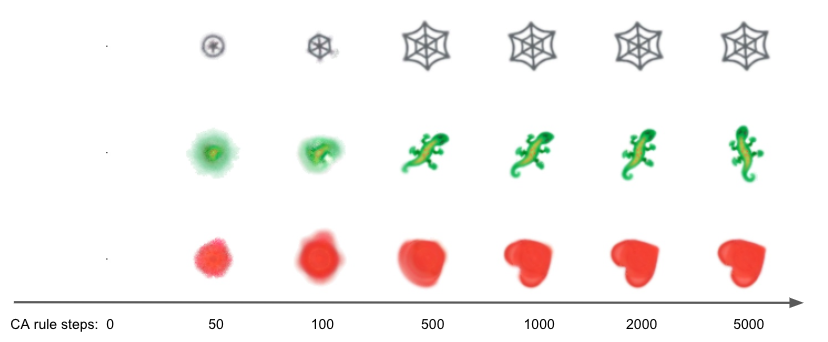}
  \caption{Unfolding of three trained Isotropic NCA rules with single seeds.}
  \label{fig:single_seed_diff_models}
\end{figure}

\begin{figure}[h]
  \includegraphics[width=\columnwidth]{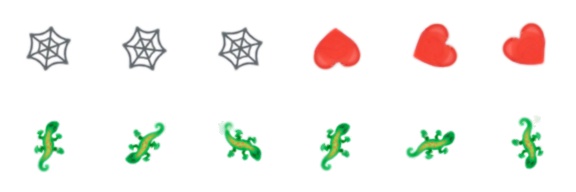}
  \caption{Step 5000 for different runs of three single seed models.}
  \label{fig:single_seed_different_runs}
\end{figure}

\begin{figure}[h]
  \includegraphics[width=\columnwidth]{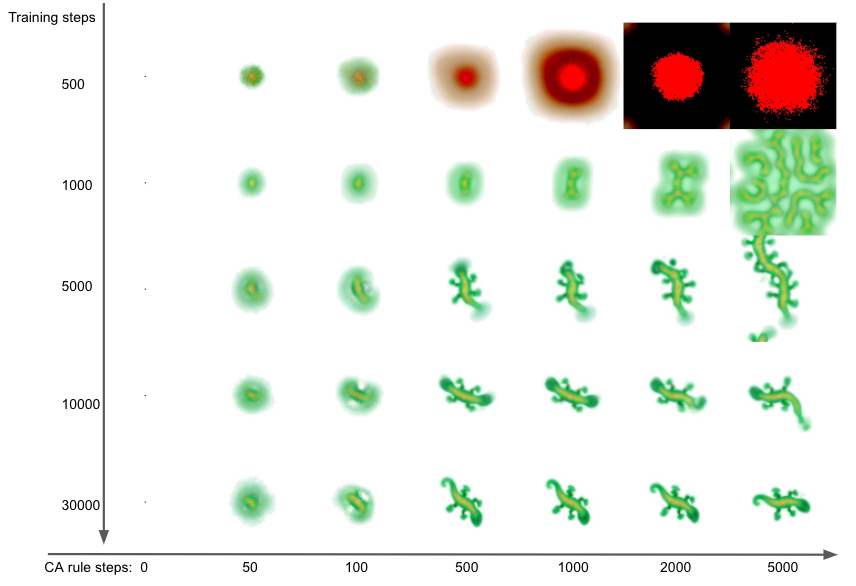}
  \caption{Evolution of the lizard IsoNCA rule during training (y-axis) versus the unfolding of each NCA rule checkpoint (x-axis).}
  \label{fig:single_seed_different_ckpt}
\end{figure}

Figure~\ref{fig:single_seed_diff_models} shows the unfolding of three different trained IsoNCA rules, starting from the original seed and running for up to 5000 steps. Here, we can observe how some patterns are more stable than others. For instance, in this training run, the lizard pattern appears to rotate over long time periods. This can sometimes happen as the rotation-reflection invariant loss used is consequently invariant to rotations over time. We observe these models to be stable for any number of steps and choose to visualize 5000 steps only because no further changes manifest beyond this point (besides rotations in the lizard case). Note in Figure~\ref{fig:single_seed_different_runs} how, for single seed experiments, every run of the same model results in a different pattern rotation and reflection.

Figure~\ref{fig:single_seed_different_ckpt} shows how the lizard model learns to form a stable pattern during training where the x-axis denotes the number of training steps (up to 5000) and the y-axis indicates the checkpoint of the model at the given training step. The model appears to first learn to generate a green, unstable blob before learning to break symmetries and, eventually, to form lizard-like features, albeit with some instability; over time, the pattern becomes more and more stable. Note how on training step 10000 the model is somewhat stable but first creates two heads and then one of them becomes a tail. Training the model for longer tends to speed up convergence to the desired form and may suppress this behaviour.

\subsection{Non-regular Grids}

\begin{figure}[h]
  \includegraphics[width=0.49\columnwidth]{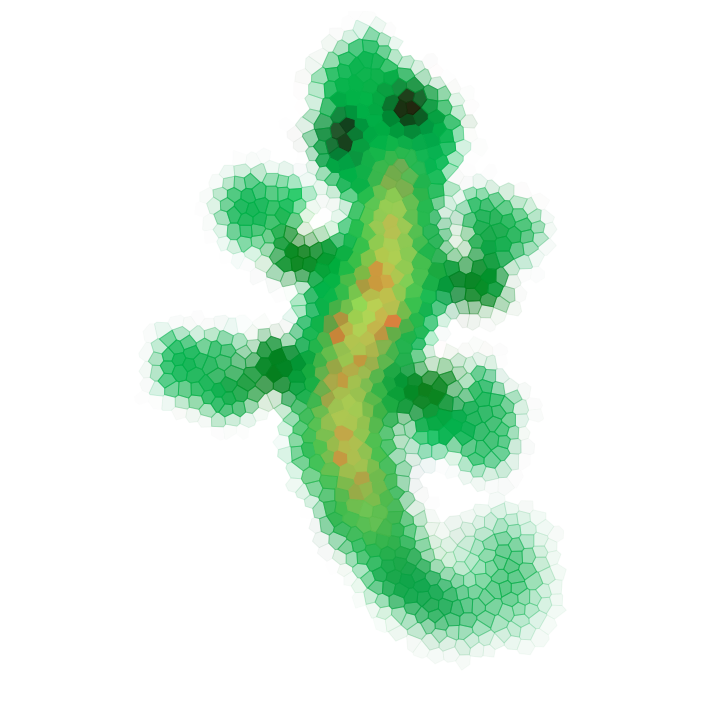}
  \includegraphics[width=0.49\columnwidth]{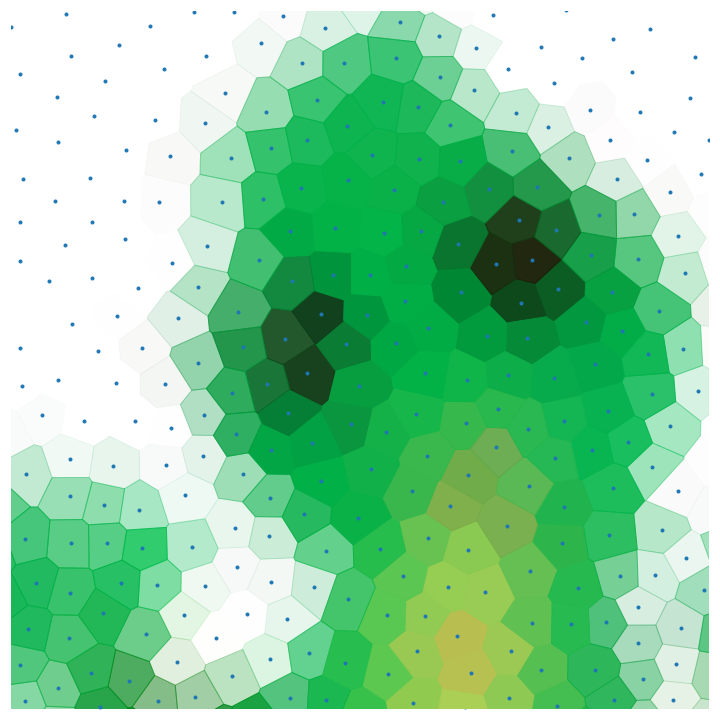}
  \caption{Example of running the lizard IsoNCA on a non-regular grid constructed using a Voronoi diagram on Poisson-disk point samples. \textbf{Left}: full pattern. \textbf{right}: close-up with voronoi cell centers shown. Running the unmodified IsoNCA rule on a non-regular grid is more prone to artifacts and instabilities. For example, in this case, the tail curl direction doesn't match that of the target pattern. }
  \label{fig:liz_voronoi}
\end{figure}

Previous work discusses the robustness of NCA models to out-of-training grid structures \citep{niklasson2021self-organising}. For example, it was found that  models trained on a square grid can be executed on hexagonal grids if 3x3 convolutional perception filters are replaced with their hexagonal counterparts. Recently, it was also demonstrated that neural models relying on a diffusion operation for communication are effectively discretization agnostic \citep{Sharp2022-ek}. In this work, we demonstrate that IsoNCA can be effectively transferred from a regular square grid to a non-regular one. To build a non-regular grid, we sample a set of points on a plane using a fast Poisson-disk sampling algorithm \citep{Bridson2007-jl}. Cells are then constructed by computing a Voronoi diagram of the sampled points. The Laplacian operator is defined as the difference between a cell's own state and the weighted average of the state vectors of its adjacent cells: $$w_{i,j} = l_{i,j} / \sum_j l_{i,j}$$ where $l_{i,j}$ is the length of the Voronoi diagram edge shared by cells $i$ and $j$ (equal to zero if there is no such edge). Figure \ref{fig:liz_voronoi} shows an example of the lizard pattern grown on a non-regular grid substrate by a square-grid-trained model using the modified Laplacian operator.

\section{Conclusion and future work}

In this work we demonstrated the capability of fully isotropic NCA models in reliably growing complex asymmetric patterns even when the perception field of each cell is fully symmetric. We believe this creates an important practical lower bound on the requirements of cell communication capabilities for morphogenesis simulations. Moreover, we think that coupling IsoNCA with physically grounded models of cell division and migration may elicit exciting possibilities for the accurate reproduction of body growth and regeneration phenomena and even enable new bio-engineering applications in the future.

\footnotesize
\bibliographystyle{apalike}
\bibliography{example} 

\end{document}